\documentclass[tinyml]{acmart}
\usepackage{xcolor}
\usepackage{graphicx}
\usepackage[caption=false,font=normalsize,labelfont=sf,textfont=sf]{subfig}
\usepackage{algorithm,amsmath}
\usepackage{algpseudocode}
\usepackage{caption}
\usepackage{soul}

\definecolor{darkgreen}{rgb}{0.0, 0.5, 0.0}
\newcommand{\sm}[1]{\textcolor{black}{#1}}

\newcommand{\short}[1]{\textcolor{black}{#1}}

\AtBeginDocument{%
  \providecommand\BibTeX{{%
    \normalfont B\kern-0.5em{\scshape i\kern-0.25em b}\kern-0.8em\TeX}}}

\setcopyright{rightsretained}
\copyrightyear{2024}
\acmYear{2024}

\begin{document}
\newcommand{\adrian}[1]{\textcolor{black}{#1}}

\title{Wet TinyML: Chemical Neural Network Using Gene Regulation and Cell Plasticity}

\author{Samitha Somathilaka}
\email{samitha.somathilaka@waltoninstitute.ie}
\orcid{0000-0003-4675-5698}
\affiliation{%
  \institution{Walton Institute, SETU, Ireland}
  \institution{University of Nebraska-Lincoln, USA}
}

\author{Adrian Ratwatte}
\affiliation{%
  \institution{University of Nebraska-Lincoln}
  \city{Lincoln}
  \state{Nebraska}
  \country{USA}
}
\email{aratwatte2@huskers.unl.edu}


\author{Sasitharan Balasubramaniam}
\affiliation{%
 \institution{University of Nebraska-Lincoln}
  \city{Lincoln}
  \state{Nebraska}
  \country{USA}
 }
 \email{sasi@unl.edu}

\author{Mehmet Can Vuran}
\affiliation{%
 \institution{University of Nebraska-Lincoln}
  \city{Lincoln}
  \state{Nebraska}
  \country{USA}
 }
\email{mcv@unl.edu}
 
\author{Witawas Srisa-an}
\affiliation{%
 \institution{University of Nebraska-Lincoln}
  \city{Lincoln}
  \state{Nebraska}
  \country{USA}
 }
\email{witty@cse.unl.edu}

\author{Pietro Liò}
\affiliation{%
  \institution{University of Cambridge}
  \city{Cambridge}
  \country{UK}}
\email{Pietro.Lio@cl.cam.ac.uk}

\renewcommand{\shortauthors}{Somathilaka et al.}

\begin{abstract}
In our earlier work, we introduced the concept of Gene Regulatory Neural Network (GRNN), which utilizes natural neural network-like structures inherent in biological cells to perform computing tasks using chemical inputs. We define this form of chemical-based neural network as Wet TinyML. The GRNN structures are based on the gene regulatory network and have weights associated with each link based on the estimated interactions between the genes. The GRNNs can be used for conventional computing by employing an application-based search process similar to the Network Architecture Search. This study advances this concept by incorporating cell plasticity, to further exploit natural cell's adaptability, in order to diversify the GRNN search that can match larger spectrum as well as dynamic computing tasks.   
As an example application, we show that through the directed cell plasticity, we can extract the mathematical regression evolution enabling it to match to dynamic system applications. We also conduct energy analysis by comparing the chemical energy of the GRNN to its silicon counterpart, where this analysis includes both artificial neural network algorithms executed on von Neumann architecture as well as neuromorphic processors. The concept of Wet TinyML can pave the way for the new emergence of chemical-based, energy-efficient and miniature Biological AI.\vspace{-0.8em}

\end{abstract}

\keywords{Biological AI, Cell Plasticity, Biocomputing, Neuromorphic.}

\maketitle

\section{Introduction}
\label{sec:Intro}

\short{TinyML aims to execute machine learning algorithms with minimum size to conserve energy consumption and deployment into devices with limited computational capacity  \cite{AnEmpiricalTinyML}. Achievements in TinyML include implementation of code sizes down to 1 KB \cite{Kusupati2018FastGRNNAF} and energy consumption of the algorithm as low as 25 mW \cite{10177729}, enabling deployment in miniature devices  such as in-body implantables\cite{basaklar2022tinyman}}.
\short{However, embedding TinyML codes in environments that cannot accomodate silicon-based devices poses a challenge, necessitating a new design paradigm that adheres to TinyML's goals of low-energy consumption and compact coding. This paper tackles this issue by shifting the focus from silicon technologies to exploring natural processes that mimics Artificial Neural Network (ANN) functions in biological cells, where we can use this to perform conventional computing  tasks.}


Our prior research delved into the computational aspect of biological cells, demonstrating that Gene Regulatory Networks (GRNs) serve as fundamental computational entities within cells, aiding in decision-making processes in response to environmental cues \cite{somathilaka2023revealing}. This entails the reception of extracellular molecules, their processing via GRNs, and the subsequent production of output molecules such as proteins. 
Nonetheless, our examination was confined to the static behaviour of cells, which does not reflect their natural adaptability.

However, further investigation on natural cell adaptability along with their learning capabilities, have prompted the question: \emph{"how can non-neuronal organisms that display traits of intelligence through plasticity be used to develop non-silicon-based neural networks?"}.
In turn, this study extends our previous work towards a new concept of \emph{Wet TinyML} that is constructed from the gene regulation process and explore the impact of cell plasticity on GRN based computing. We refer to this component of Wet TinyML as {\bf Gene Regulatory Neural Networks} (GRNNs), where we transform a GRN with established weights based on the relative gene expression. In the GRNN, genes can receive molecules known as Transcription Factors (TF) from multiple other neighboring genes \cite{Hao2011}, akin to how perceptrons receive external inputs. The influence of one gene on another acts as the 'weight' in this analogy. The cumulative impact of molecular signals from neighboring genes and their intensities collectively regulate the expression of a target gene, analogous to the weighted summation in ANN processing. Furthermore, a gene is expressed only if the cumulative influence of the transcription factors exceeds a certain threshold \cite{Spitz2012}, mirroring the role of activation functions in ANNs. In a number of ways, the GRNN can be associated to a chemical-hardware version of a neuromorphic computing system.
\begin{figure}
    \centering
    \includegraphics[trim={0 0 0 0},clip,width=0.9\linewidth]{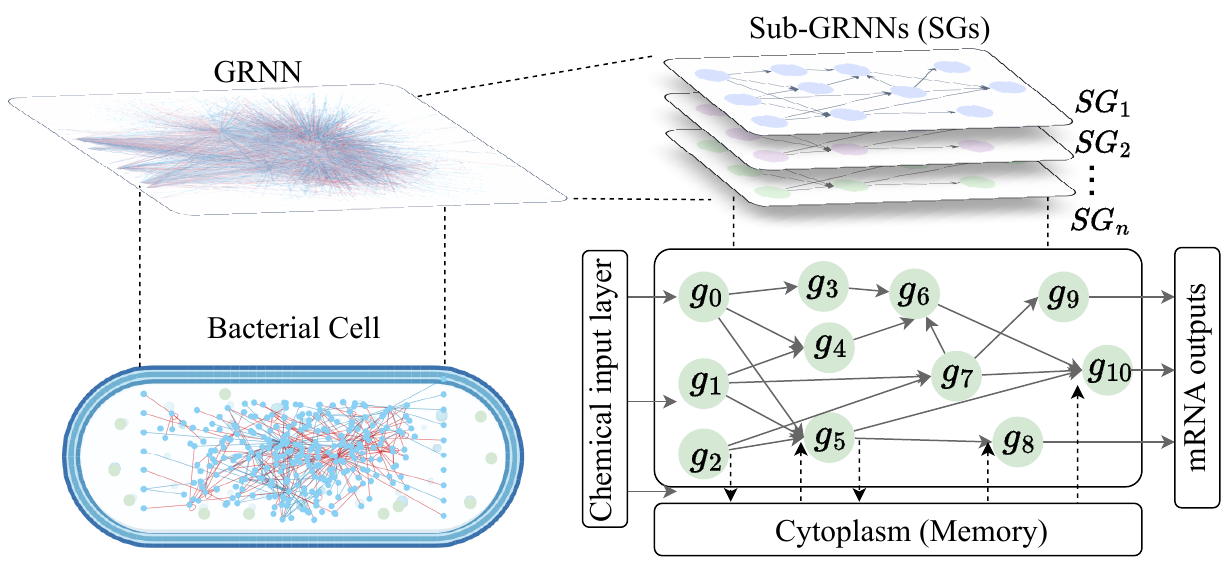}
    \caption{Wet TinyML based on GRNNs extracted and searched in a bacterium to perform computing. Input chemicals trigger selective activation of relevant GRNN-subnetworks, rendering the GRNN a composite of many subnetworks.  Gene products, diffusing into the cytoplasm forming cellular memory system that contributes to cell plasticity. \vspace{-2.4em} }
    \label{fig:Intro}
\end{figure}

\begin{figure}
    \centering
    \includegraphics[trim={11 9 30 4},clip,width=0.9\linewidth]{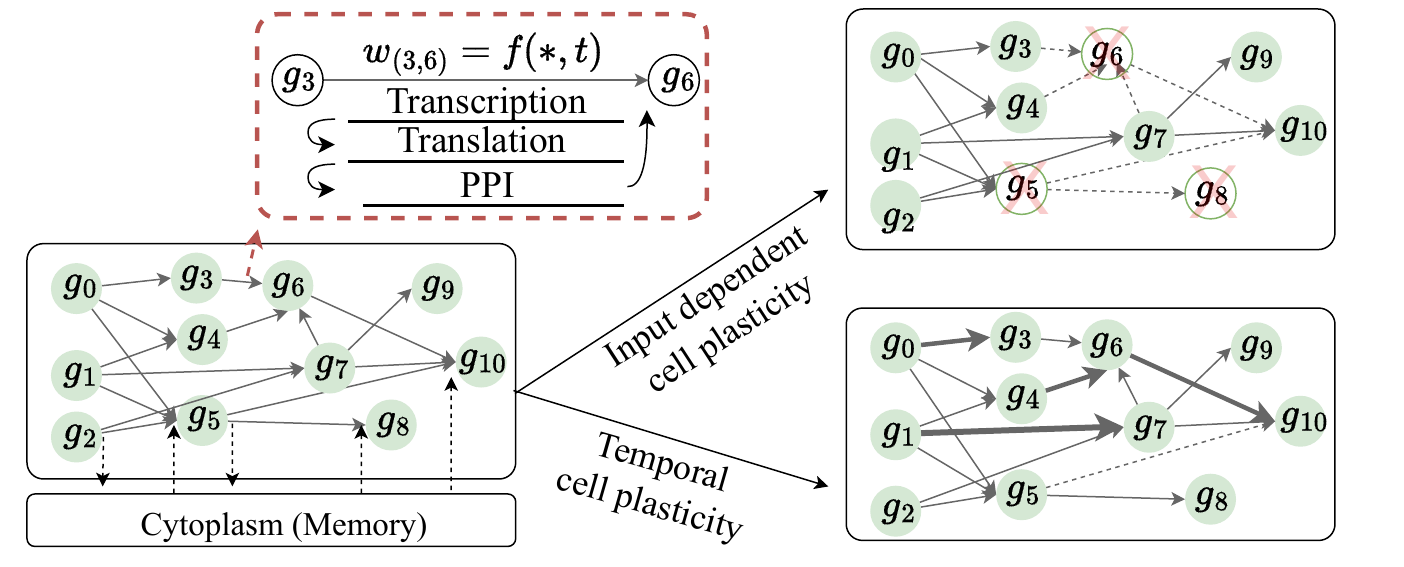}
    \caption{In GRNN framework, gene-perceptron operate similarly to ANN perceptrons, processing inputs with weights influenced by multi-omic layer interactions \sm{(*), and time(t)}. Bacterial cells exhibit input-dependent plasticity by unique gene expression pathways varying with different inputs. Additionally, cells demonstrate temporal plasticity by altering GRNN subnetwork interaction weights over time.}\vspace{-1.5em}
    \label{fig:Plasticity}
\end{figure}

Research indicates that gene expression is selectively influenced by input chemicals, suggesting that the GRNN can be viewed as a large collection of pre-trained GRNN sub-networks and each is triggered depending on its chemical input such as nutrient molecules (Fig. ~\ref{fig:Intro}). Each GRNN subnetwork comprises an input layer, intermediate nodes that are akin to the hidden layer, and an output layer. The products of the transcription and translation processes resulting from natural computing are diffused into the cytoplasm as depicted in Fig. ~\ref{fig:Intro} and interact with other biological components \sm{such as ribosomes} \sm{\cite{Rodnina2018}}. Further, as Fig. ~\ref{fig:Plasticity} elucidates, \sm{the accumulated molecules in the cytoplasm can act as a memory module that induces feedforward and feedback signals in optimizing GRNN subnetwork switching and adjusting weights} over time.


The GRNN, as a pre-trained network, allows bypassing the conventional ANN training phase by directly selecting an appropriate GRNN sub-network for specific tasks\cite{Somathilaka2023}, similar to the way Network Architecture Search (NAS)\cite{adam2019understanding} using supervised data. Subsequently, this research examines expanding this GRNN sub-network search space by accounting cellular plasticity and assesses energy consumption compared to existing neuromorphic systems. Finally, the study utilizes GRNN-subnetworks to derive various mathematical regression models.

This paper is organized as follows: Section \ref{sec:wetTinyML} introduces 'Wet tinyML', explaining the inherent computing power of natural biological cells driven by GRNNs. Section \ref{sec:Analsys} compares GRNN's energy efficiency with traditional von Neumann and neuromorphic systems, focusing on algorithmic and structural complexities, and discusses how natural cell plasticity can enhance computing diversity. The applications of GRNN are detailed in Section \ref{sec:Application}, and Section \ref{sec:conclusion} concludes the study.\vspace{-1em}



\section{Background of GRNN and Cell Plasticity}
\label{sec:wetTinyML}

This section introduces the elements of GRNN for Wet TinyML, building upon our earlier research \cite{somathilaka2023revealing, Somathilaka2023}.\vspace{-1em}
\subsection{Gene-perceptrons and Weights}

As discussed in Section \ref{sec:Intro},  several characteristics of genes in their complex regulatory process exhibit similarity to an artificial neuron. 
In ANNs, a perceptron processes multiple inputs by applying weights and summing them, followed by an activation function. This process finds parallels to genetic circuit operations, where a gene receives TF molecules that leads to a combinatorial regulation of a gene's expression. 

In ANNs, the output from the weighted summation is modulated by activation functions, such as \emph{sigmoid}, \emph{tanh}, or \emph{Rectified Linear Unit} (\emph{ReLU}), which introduces non-linearity to the computing. This concept is mirrored in gene regulation, where a gene can be in an `on' or `off' state, depending on the regulatory impact of the TFs. The combined influence of TFs acts like the weighted summation, translating into binary gene expression states (`on' or `off'), very similar to a sigmoid function's output.

However, when considering the time dynamics of gene expressions, we showed in our previous work \cite{Somathilaka2023} the ReLU activation function is more compatible than the sigmoid function, as it accommodates the linear relationship between the TF influence and gene expressions over time.
Further observation reveals that prokaryotic genes exhibit `ground states', where RNA polymerase can bind to promoters in the absence of activators or repressors, suggesting an addition of a bias term to the weighted summation of each gene. Given these gene properties that behave as an ANN perceptron, we introduced the term \textbf{gene-perceptron} in the GRNN \cite{somathilaka2023revealing, Somathilaka2023}.\vspace{-1em}

\subsection{GRN-to-GRNN Conversion}
\label{sec:GRNToGRNN}

The GRNN is considered a pre-trained graph-strucured neural network that is inferred by assigning weights to gene-gene interactions of a GRN, based on relative gene expression levels, which we detail in this section.
The weight extraction for each gene-perceptron is conducted iteratively using network modules comprising the target gene-perceptron and its associated source gene perceptrons, structurally akin to single-layer perceptrons. 

\begin{figure}
    \centering
    \includegraphics[trim={22 5 65 0},clip,width=0.9\linewidth]{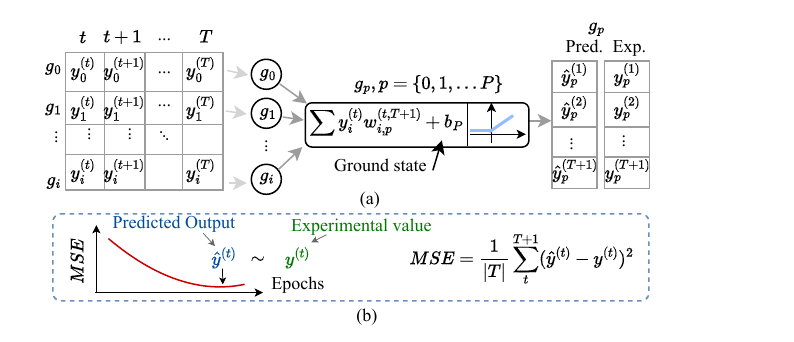}\vspace{-1.5em}
    \caption{Illustration of the weight extraction process where a) elucidates the utilization of transcriptomic data considering gene $g_P$ as a target single-layer gene-preceptron and b) depicts the training process of minimizing the MSE between predicted and experiment expression levels.} \vspace{-1.5em}
    \label{fig:WeightExtraction}
\end{figure}

The weight extraction for each gene-perceptron module involves a process similar to training a single-layer perceptron and involves the use of transcriptomic data as shown in Fig. \ref{fig:WeightExtraction}a. In this figure, $y^{(t)}_{i}$ is the expression level of gene $g_i$ at timestep $t$, $w^{(t,T+1)}_{i,p}$ is the weight of the interaction between gene $g_i$ and $g_p$ in the time interval $t$ to $T+1$, and $b_P$ is the bias (the ground state) of the gene $g_p$. Initially, random weights are introduced with the experimental transcription data at the levels of source gene-perceptrons. The target gene-perceptron's predicted transcription level is then calculated by passing the weighted summation of experimental source gene-percepton levels and weights through a ReLU function. The weights are then refined based on adjusting the differences between the predicted and experimental transcription levels of the target gene-perceptron by using learning rate as 0.001 with $10^5$ epochs. This iterative adjustment continues across all experimental transcriptomic data until the error is minimized as shown in Fig. \ref{fig:WeightExtraction}b. Repeating this procedure estimates weights for all gene-perceptron modules, which are then applied to the GRN, transforming into a GRNN. More details on this weight extraction can be found in \cite{Somathilaka2023}.\vspace{-0.8em}

\subsection{Input-dependent and Temporal Plasticity}
Bacteria are renowned for their remarkable adaptability in various environments, a trait crucial for their survival \cite{zhang2022selective}. In observing bacterial cells as natural computational entities, we analyze this cellular plasticity through the lens of ANNs, identifying two primary features: input-dependent plasticity and temporal plasticity. 

\textit{Input-dependent plasticity} is driven by the selective responsiveness of genes to specific input chemicals \cite{zhang2022selective}. Genes at the periphery of the GRNN, akin to the input layer in an ANN, are particularly sensitive to certain chemical effectors. In nature, this sensitivity leads to the expression of a specific subset of genes in response to the abundance of these chemical molecules. The GRNN, thus, selectively channels information flow, activating only the relevant expression pathways, while keeping other genes largely idle. This results in the utilization of specific GRNN subnetworks based on the chemical input, as illustrated in Fig.~\ref{fig:Plasticity}. This in turn enhances energy efficiency during the gene regulatory process of the cells.

On the other hand, \textit{temporal plasticity} involves alterations in the influence of one gene on another over time \cite{rivera2021framework} to achieve the optimal behavior for an given environment, which resembles weight plasticity within the same input conditions. 

In previous works \cite{somathilaka2023revealing, Somathilaka2023}, we focused on a static GRNN to search application-specific GRNN subnetworks. However, this study focuses on harnessing these natural plasticities to expand the diversity of searching for the optimal applications specifc GRNN subnetwork. \vspace{-1em}

\section{Energy and Computing Dynamics of GRNN}
\label{sec:Analsys}

\subsection{Energy vs Computing Complexity}

The scale and energy efficiency of ANNs are critical factors in the miniaturization of algorithms and neuromorphic hardware designs. Biological systems, such as cells and neurons, which operate at the micrometer scale, demonstrate exceptional energy efficiency in natural computing processes. For example, the human brain, despite its immense computational capacity, has a remarkably low energy consumption of only 20W \cite{gebicke2023computational}. Similarly, the expression of a single gene-perceptron, a complex task in itself, consumes a mere 0.01 fW \cite{lane2010energetics}. This observation leads to the consideration of the energy profile of GRNNs as a potential full-scale computing platform.

\subsubsection{GRNN Structural and Algorithmic Complexity}
Taking into account that the GRNN is a randomly structured network with power-law degree distributions, we will analyze the estimated algorithmic and structural complexities. The algorithmic complexity reflects the information diffusion, failure propagation, and resilience through the network \cite{morzy2017measuring}. It is defined by the Kolmogorov complexity, which is approximated using the Coding Theorem Method \cite{zenil2018decomposition}. Understanding the algorithmic complexity, which reflects on the complex nature of the network structure, can provide avenues for interpretability of the model \cite{9380482}. The structural complexity, on the other hand, is determined by the betweenness centrality and relative degree of gene-perceptrons in the network \cite{ghanbari2023structure}. Although our previous work analyzed the structural and algorithmic complexity in GRNNs \cite{Somathilaka2023}, in this paper we use these two measures to determine the relationship to the energy consumption of the GRNN. 
Further information regarding the use of the structural and algorithmic complexiy in GRNNs can be found in \cite{Somathilaka2023}.

\subsubsection{GRNN Energy Analysis}
We will now determine how the algorithm and structural complexities of the GRNN and comparison to ANN play a role on the energy consumption. The total energy consumption for the \(i^{th}\) GRNN, denoted as \(P_{total}(i)\), is computed by summing the energy used in the transcription and translation processes, given by \(P_{ex}(i) + P_{tra}(i)\). Here, \(P_{ex}(i)\) represents the peak power for gene expression in the \(i^{th}\) GRNN, and \(P_{tra}(i)\) is the power required for its translation process. The transcription power is calculated as \(P_{ex}(i) = |GRNN_i| \cdot \hat{p}_{ex}\), where \(\hat{p}_{ex}\) is the per gene-perceptron transcription power and \(|GRNN_i|\) is the number of gene-perceptrons in the \(i^{th}\) GRNN. As mentioned earlier, $\hat{p}_{ex} = 0.01$ fW for a medium-sized prokaryotic cell such as \textit{Escherichia coli} \cite{lane2010energetics}. However, studies show that approximately 75\% of a cell's energy dissipation is attributed to translation processes, while a mere 2\% is utilized for transcription \cite{harold1987vital}. Subsequently, using this 2:75 power ratio between $P_{tra}$ and $P_{ex}$, the total power $P_{total}(i)$ is calculated.

In parallel, the energy consumption of silicon-based computing units is calculated as $P_{total} = N \cdot \hat{p}$, where $N$ is the number of neurons in the system and $p$ is the unit power. According to literature, the unit power consumption of a neuron for different processors, $p$, are listed as follows: Spikey at 1.49x$10^{-06}$, R2600X at 9.62x$10^{-04}$, Intel mobile at 3.37x$10^{-04}$, and RTX2070 at 3.18x$10^{-05}$ \cite{ostrau2022benchmarking}.

\begin{figure}
    \centering
    \includegraphics[trim={45 0 20 0},clip,width=0.90\linewidth]{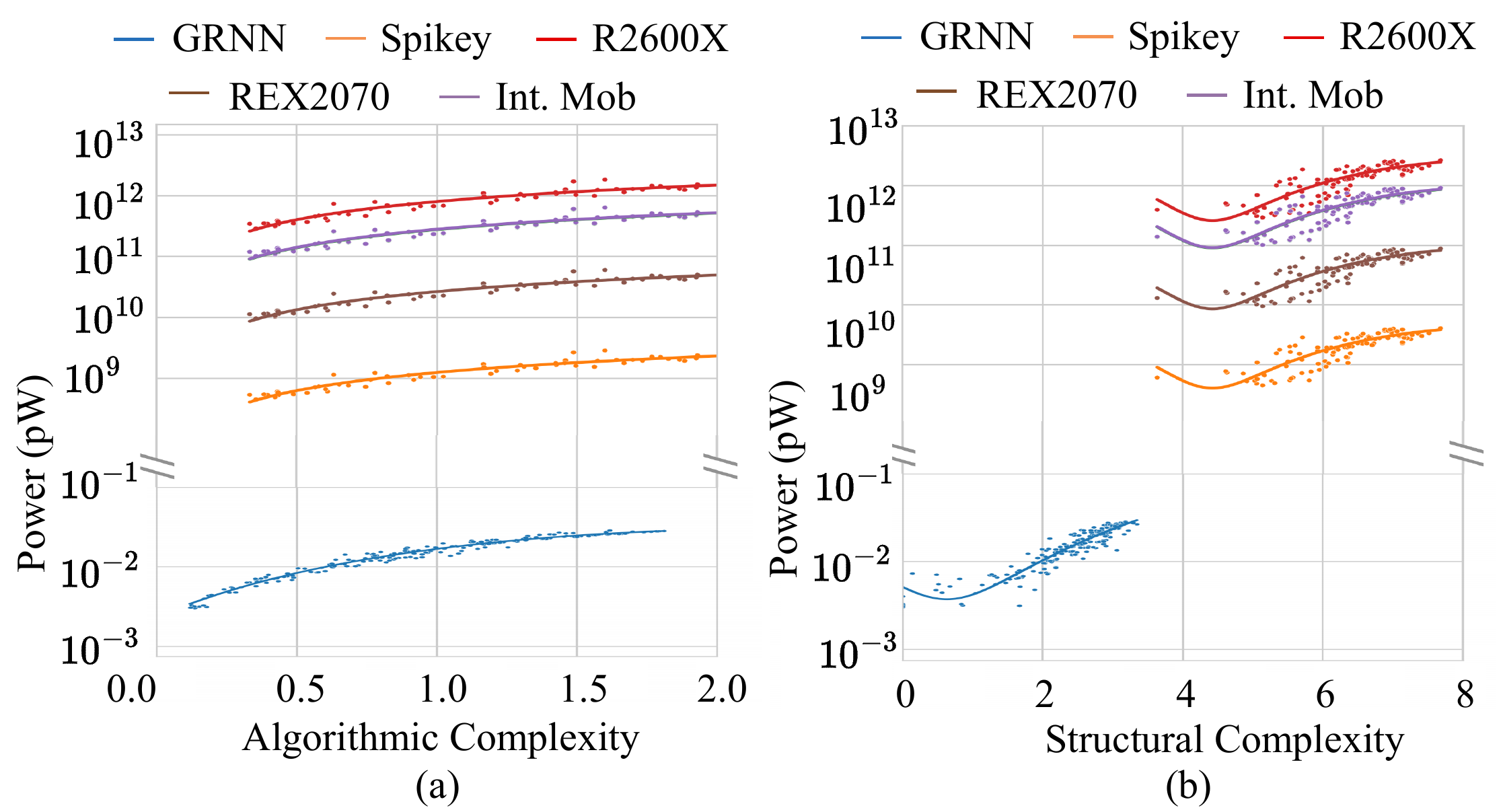}\vspace{-.8em}
    \caption{Power comparison between GRNN vs von Neumann and neuromorphic computing systems with respect to a) algorithmic complexity and b) structural complexity. \vspace{-1.5em}}
\label{fig:EnergyComparison}
\end{figure}

We emphasize that the energy consumption across various computing platform focuses solely on the energy used for computing, excluding housekeeping energy requirements. 
We first compare the energy consumption of GRNN with four other processors mentioned above with respect to the algorithmic complexity of 200 different model sizes. This evaluation involved varying the number of nodes within each model across four Von Neumann and neuromorphic platforms. 
The results of this analysis are presented in Fig.~\ref{fig:EnergyComparison}. Notably, the GRNN's maximum power consumption does not surpass 0.05 pW, even at the highest level of algorithmic complexity, as shown in Fig.~\ref{fig:EnergyComparison}a. In contrast, the other platforms register energy usage ranging from $10^9$ pW to $10^{12}$ pW for models with an equivalent number of neurons. 
The less sparse connectivity and low diameter in the GRNN typically contribute towards minimized algorithmic complexity. While the  heterogeneity in the weights can increase the algorithmic complexity, this increase still results in low energy consumption due to the chemical energy used by the gene-perceptrons \cite{Zenil2018}.

Furthermore, a similar comparison is conducted to examine energy dissipation with respect to the structural complexity, depicted in Fig.~\ref{fig:EnergyComparison}b. The results uncover patterns of energy consumption similar to those observed in the energy dissipation with respect to the algorithmic complexity. It is important to highlight that the small-world network structure of the GRNN exhibits notably lower structural complexity compared to others. Small-world networks, known for their high clustering and short path lengths, have a more orderly structure, which lowers their structural entropy \cite{freitas2019detailed, omar2020survey} and this can be observed in Fig.~\ref{fig:EnergyComparison}b. 

The additional housekeeping energy of GRNN computing depends on a range of factors including administration of chemical inputs and extraction method of outputs, which will be explored in our future research.\vspace{-0.8em}


\subsection{Non-Plasticity \sm{Search Space} of GRNNs}
\begin{figure}
    \centering
    \includegraphics[trim={45 5 0 0},clip,width=0.9\linewidth]{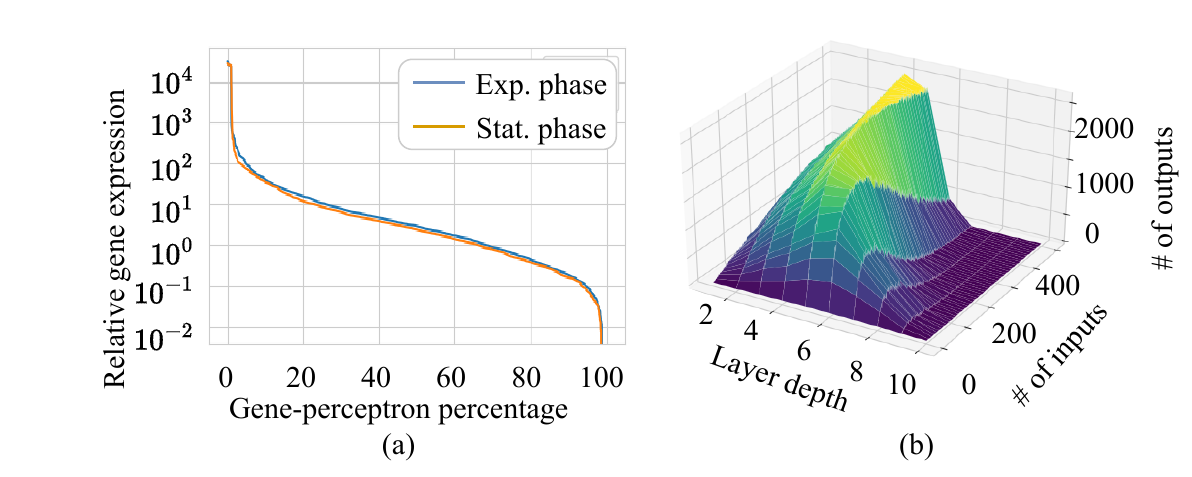}
    \caption{Illustrations of a) the sparsity of gene expression and b) number of output node variations given the number of input nodes and the depth of the GRNN subnetwork.\vspace{-1.6em}}
    \label{fig:SparcityAndInOut}
\end{figure}

This section first assesses the sparsity of GRNN by analyzing gene expression levels during two key growth stages: exponential growth and the stationary phase, using data sourced from \cite{sutormin2022interaction}. As illustrated in Fig.~\ref{fig:SparcityAndInOut}a, on average, a cell utilizes only about 10\% of its genes at the considered two growth phases of bacteria, which are the idling and rapid growing phases, respectively. This observation indicates that the cell possesses a broader range of options for transitioning between GRNN subnetworks to adapt to diverse environmental conditions. The inherent sparsity in the GRNNs further contributes significantly to massively parallelized computing. This explains how bacteria in nature process signals in a parallel manner, regulating multiple metabolic pathways simultaneously. Such natural parallels further underscore the capacity of GRNNs for parallel computing.

\short{We investigate the diversity of GRNN subnetworks by focusing on the size of input, intermediate hidden layer, and output gene-perceptrons. By selecting 100 random input gene-perceptrons and tracking their connections through the network, we analyzed the structure up to 10 layer depths. This process is repeated 100 times with varying input sets aiming to average the count of gene-perceptrons per layer. The experiment is conducted with input sets increasing by 100, up to 500 gene-perceptrons, to ensure a comprehensive assessment.}

It is evident from Fig.~\ref{fig:SparcityAndInOut}b that with a relatively small sized  network with the input layer comprising 100 nodes, which is capable of processing inputs for 100 features, the maximum output node count reaches approximately 500 when the network depth is close to 6. This GRNN subnetwork diversity allows the selection of a certain number of output nodes, up to 500, tailored for a given application. For instance, in an application requiring 10 output nodes, the GRNN offers approximately $8.9 \times 10^{26}$ combinations of output node choices. It is important to note that this abundance of options results from the initial choice of 100 nodes in the input layer. Considering only 1000 suitable candidates for the input layer, the permutations for 100 input nodes increases to $5.9 \times 10^{297}$. This astronomically large number of combinations underscores the GRNN's possibility of facilitating the identification of a suitable GRNN subnetwork for specific applications exhibiting generalizability.

Furthermore, Fig.~\ref{fig:SparcityAndInOut}b
reveals that increasing the number of nodes in the input layer to 500 results in the expansion of the output layer to approximately 2,500 nodes. This provides $9.3 \times 10^{33}$ options for applications requiring 10 outputs, when the network depth is around six. This demonstrates the significant impact of input layer size on the diversity and adaptability for the network's output. 

\subsection{Search Space Expansion with Cell Plasticity}


\adrian{ In this section, we examine input-dependent and temporal cell plasticity using the weight extraction algorithm from our previous study \cite{somathilaka2023revealing}, and data sourced from the GEO database \cite{barrett2012ncbi} to evaluate their impact on gene expression within the GRNN.} 
\subsubsection{\adrian{Input-dependent Cell Plasticity}}
\begin{figure}[t]
\centering
\subfloat[\label{fig:shortest_dist}]{
\includegraphics[trim={0 0 0 15}, clip, width=0.40\linewidth]{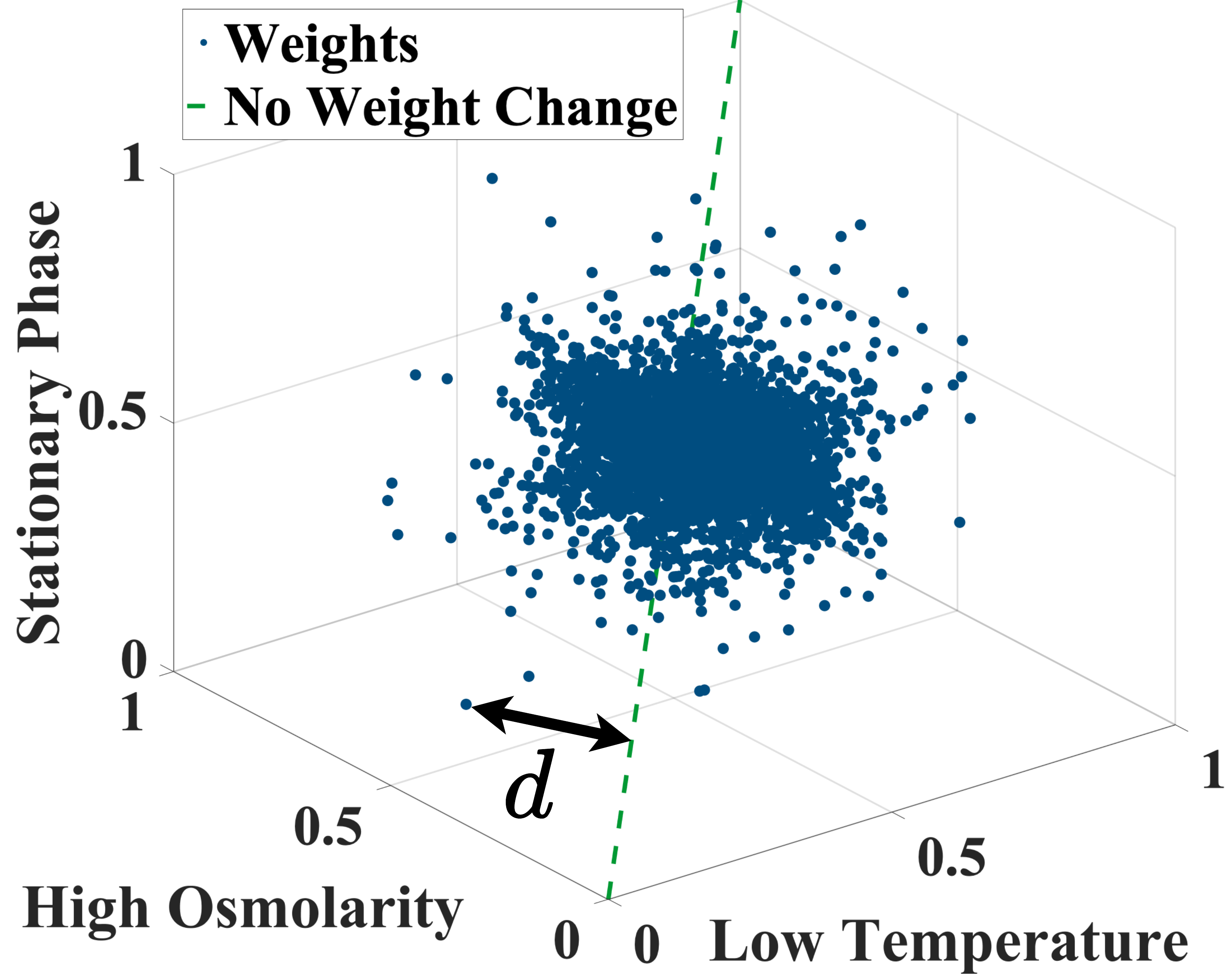}}
\subfloat[\label{fig:condition_plas}]{
\includegraphics[trim={0 0 0 10},clip,width=0.40\linewidth]{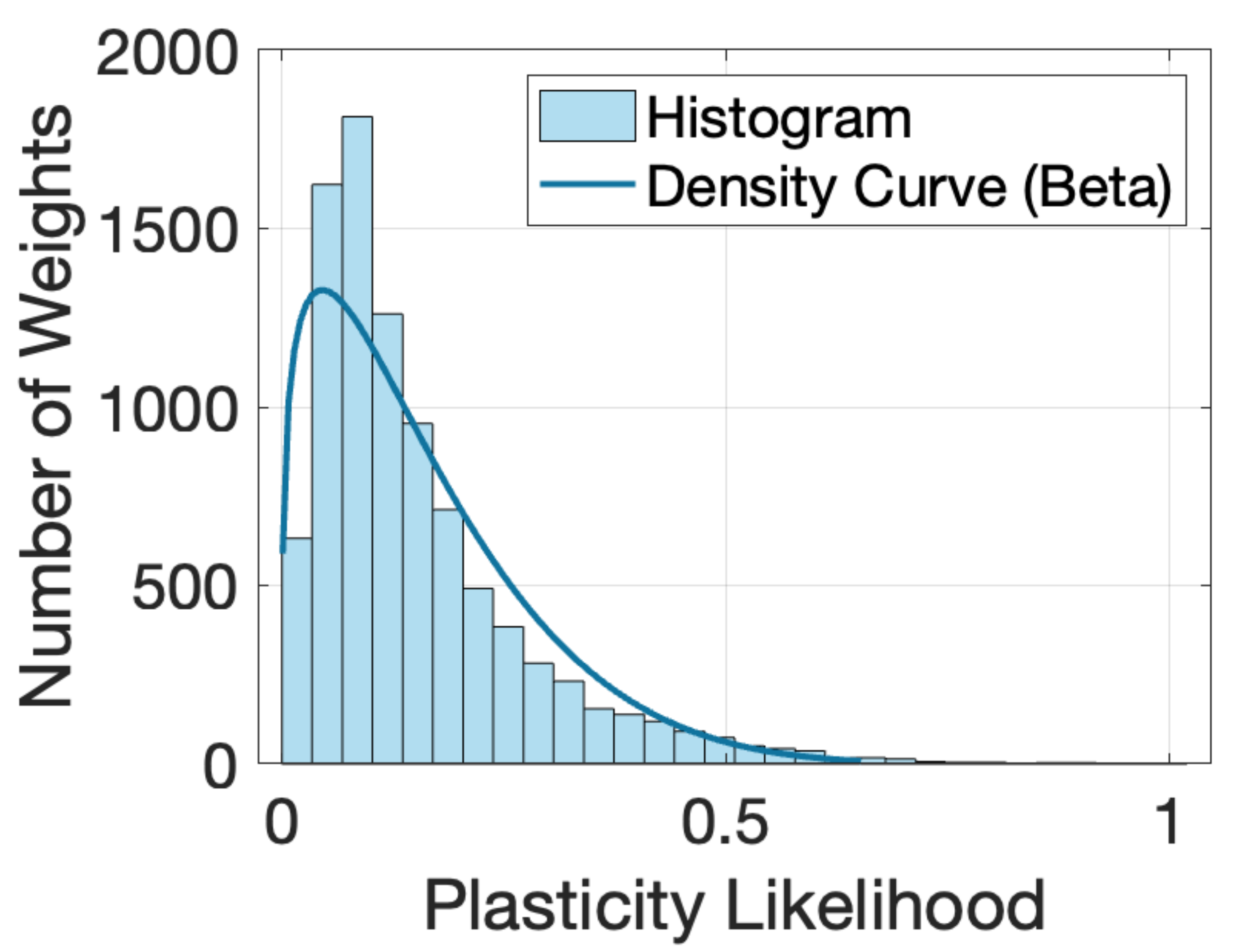}} \vspace{-1em}
\caption{Environmental Conditions and Cellular Plasticity: (a) Weight changes under conditions of low temperature, stationary phase, and high osmolarity, with $d$ indicating the distance from the no weight change line. (b) Plasticity likelihood related to weight adjustments across various inputs. \vspace{-1.5em}}
\end{figure}

\short{We study input-dependent cell plasticity using transcriptomic time series data from three environmental settings: low temperature, high osmolarity, and stationary phase. Weight changes across inputs are calculated by determining the geometric shortest distance between extracted weights for each condition and a line connecting (0,0,0) to (1,1,1). This line indicates no changes in the gene expression for the three conditions. Fig. \ref{fig:shortest_dist} illustrates that while some weights align with the no change line, the majority exhibit varying degrees of change. Fig. \ref{fig:condition_plas} depicts the probability of weight plasticity across all three conditions, showing a left-skewed Beta density curve indicating that most weights have probabilities less than 0.5 for undergoing changes. Furthermore, our analysis in Table \ref{tab:condition_plas} explores how individual condition changes influence weight alterations across all conditions, revealing that approximately $2-5\%$ of the total weights have changed between distinct input conditions. This pattern emerges from selective gene activation by environmental factors, rather than affecting all genes universally. This analysis showcases the GRNN's capacity to adapt weights selectively in response to new input conditions, without affecting all weights uniformly.}\vspace{-0.5em}

\subsubsection{\adrian{Temporal Cell Plasticity}}
\begin{figure}[t!]
    \centering
    \vspace{-1em}
    \includegraphics[width=0.9\linewidth]{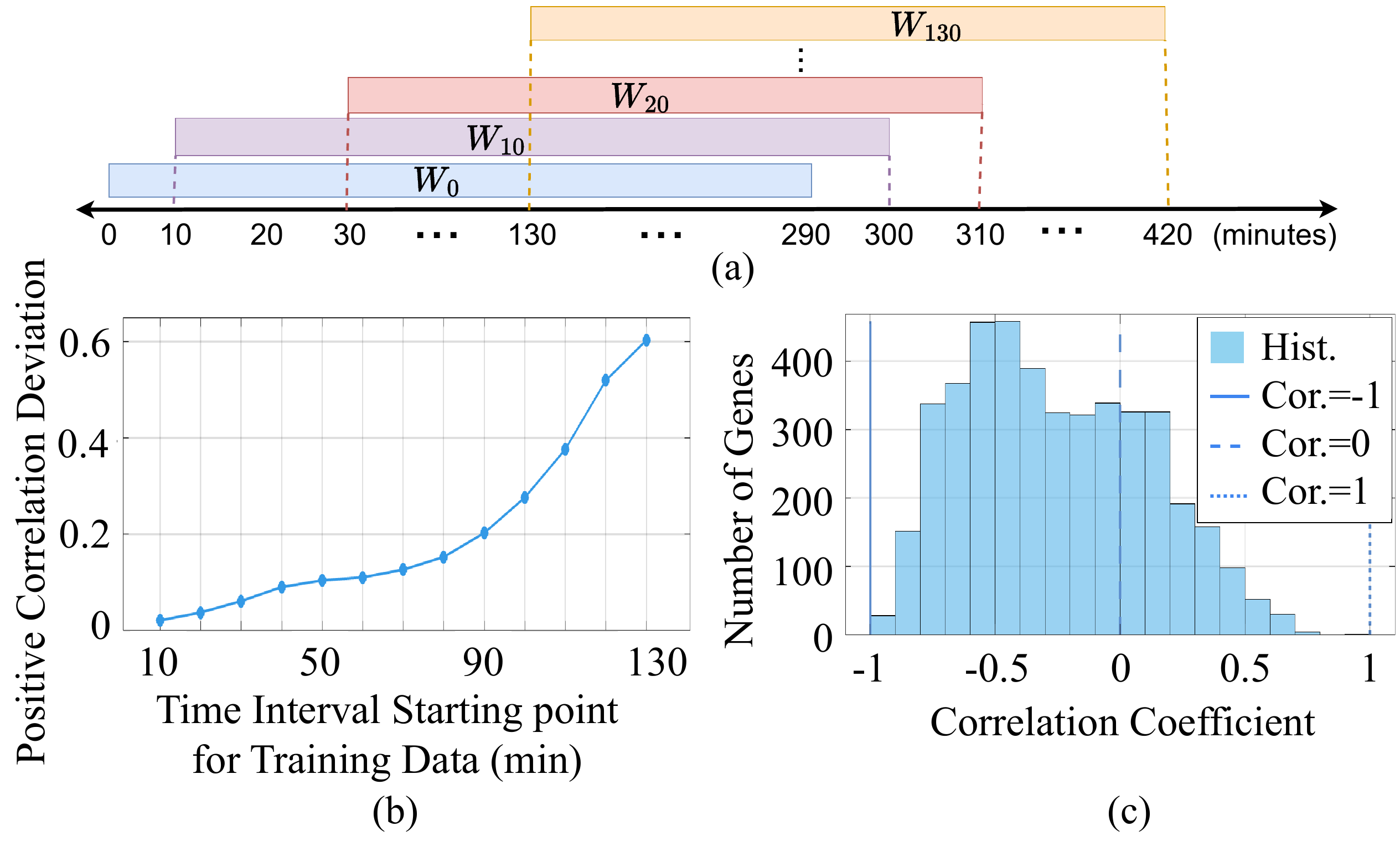} \vspace{-1.5em}
    \caption{Temporal plasticity a) data partitions,  b) deviation from positive correlation and c) correlation of gene expression between 0-290 and 130-420 minutes time intervals. \vspace{-1.5em}}
    \label{fig:chunk_deviate}
\end{figure}

In this section, we analyze the temporal weights change within the GRNN. \adrian{To analyze this, we utilize the transcriptomic data (GSE65244), which encompasses gene expression levels recorded at 10-minute intervals within the range of 0 to 420 minutes.}  We segment this dataset with respect to time into equal-sized partitions, each containing 30 expression levels collected at 30 different time points. \adrian{The weights extracted for each of these time periods are represented as $W_0$, $W_{10}$, ..., $W_{130}$, as depicted in Fig. \ref{fig:chunk_deviate}(a).}
We compute the correlation coefficient between the weights extracted for the initial time period (0-290 minutes), serving as the reference ($W_0$), and the weights from each subsequent time period.
Fig. \ref{fig:chunk_deviate}(b) illustrates the deviation of this computed correlation from the ideal positive correlation of 1. The deviation gradually rises to 0.1 between the time period 10-300 and 60-350 minutes at a gradual pace, before increasing rapidly. 
This suggests that applying identical input conditions for an extended duration prompts the GRNN to update its weights, contributing to the cell's survival as part of its plasticity process.
As a result of this deviation from the positive correlation, we attain GRNNs with different weights over time, thereby expanding the search space for the GRNN sub-networks to suit an application. Since weight changes occur primarily in specific parts of the global GRNN, we can extract sub-networks featuring both dynamic and static weights.
\adrian{ Tailoring input conditions over specific time periods allows us to derive the desired set of weights suited for the application.}\vspace{-0.5em}

\begin{table}[t]
\caption{Frequency analysis of altered weights across different experimental conditions. \vspace{-1em}}
\label{tab:condition_plas}
\resizebox{\columnwidth}{!}{%
\renewcommand{\arraystretch}{0.8}
\begin{tabular}{lll}
\hline
Exp. replicate & \begin{tabular}[c]{@{}l@{}} \adrian{\# of Weights Showing Higher} \\ \adrian{Probability of Plasticity}\end{tabular} & Ratio ($$\%$$) \\ \hline
All Conditions & 466 & 8.08 \\ \hline
Low temperature & 372 & 5.06 \\ \hline
High Osmolarity & 130 & 2.42 \\ \hline
Stationary phase & 241 & 4.63 \\ \hline
\end{tabular}
}
\end{table}

\subsubsection{Correlation Analysis of Gene Expression Temporal Dynamics in the GRNN}
The analysis of temporal gene expression dynamics aims to assess the influence of the temporal weight changes, previously addressed, on the gene expression within the GRNN. 
We compute correlation coefficients for gene expression levels between the time intervals of 0-290 and 130-420 minutes for each gene. 
Fig. \ref{fig:chunk_deviate}c presents this results and shows that \adrian{roughly $80\%$} 
of genes within the GRNN display negative correlation, with \adrian{correlation coefficients} predominantly distributed between 0 and -1, while around $20\%$ of the total genes exhibit positive correlation. 
This correlates with the results in  Fig. \ref{fig:chunk_deviate}c, where majority of weights undergo significant changes over time and this results in variations in the expression of most genes within the GRNN. \vspace{-0.7em}




\section{GRNN Application for Regression}
\label{sec:Application}

Unlike traditional ANNs, bacterial GRNNs was introduced with a specialized algorithm to search relevant subnetwork omitting the conventional ANN training, identifying input/output genes and the optimal time window based on supervised application data. This section elucidate a use case of regression to show the contribution of cell plasticity in expanding the search space theoretically. Although, the weight extraction method discussed in Section \ref{sec:GRNToGRNN} and \cite{Somathilaka2023} can be used for any temporal transcriptomic and GRN data, this study uses \textit{E. coli} as the model species.


\begin{figure}
    \centering
    \includegraphics[trim={0 0 20 12},clip,width=0.87\linewidth]{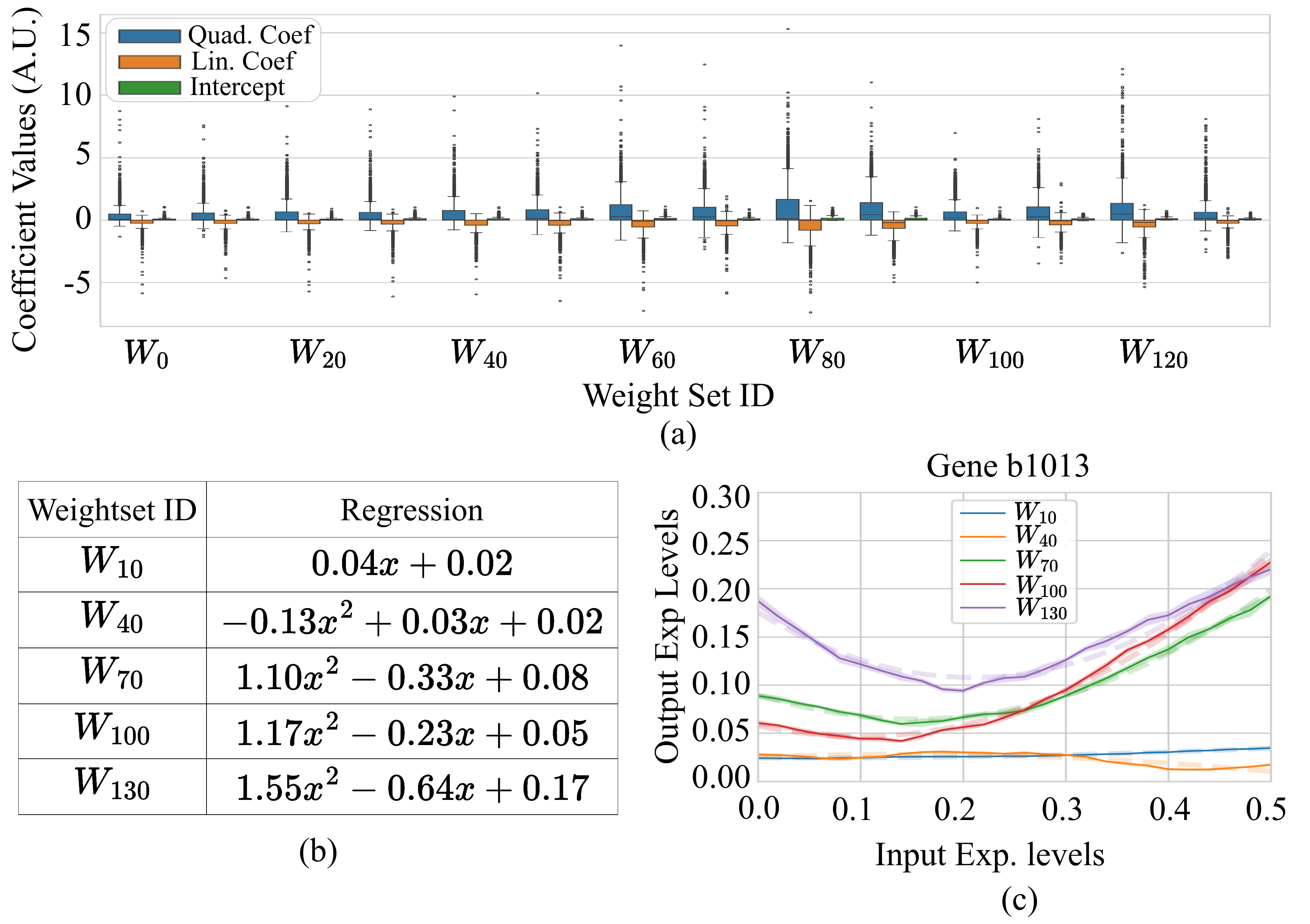} \vspace{-0.8em}
    \caption{ The regression analysis, using \textit{b3067} as the sole input gene-perceptron, includes: a) analysis of quadratic, linear coefficients, and intercepts across weight configurations outlining the solution space; b) the regression coefficients of \textit{b1013}; and c) the associated regression curves. }\vspace{-2.1em}
    \label{fig:RegressionAnalysis}
\end{figure}


This analysis utilizes seven distinct time-based weight configurations (\(\mathbf{W}_i\), where \(i=\{0, 10, 20, \ldots, 130\}\)) as shown in Fig. \ref{fig:chunk_deviate}a to assess the diversity in regression functions. To sharpen our focus, we select a single gene-perceptron as the input and focus only on quadratic regressions. Gene \textit{b3067} is chosen as the input layer gene-perceptron for its impact on 1702 gene-perceptrons to get a larger  solution space. This  gene-perceptron is stimulated with five input concentrations (0.1 to 0.5). Based on ten iterations per setup to capture the stochastic behaviour, the output gene-perceptrons' expression levels are averaged. Expression levels are recorded at each timestep in different weight setups $W_i$ and using curve fitting for quadratic functions, quadratic and linear coefficients, and intercepts of gene-perceptrons are determined.

Fig. \ref{fig:RegressionAnalysis} illustrates the regression diversity of the GRNN with respect to the temporal cell plasticity. 
For each weight configuration \(w_i\), Fig.~\ref{fig:RegressionAnalysis}a displays the variations in quadratic and linear coefficients, as well as intercepts. Each box plot in the figure represents the coefficients for 2,875 gene-perceptrons. Notably, there is a substantial variation in quadratic coefficients across all weight configurations, resulting in a range of regressions with varying curvatures. In turn, each box plot emphasizes the diversity in the solution space for a given application. Moreover, the distribution of these quadratic coefficients highlights the potential for deriving linear regressions in cases where the quadratic coefficient equals zero.
However, the linear coefficients of curves linked to the chosen input gene-perceptrons (\textit{b3067}) tend to exhibit predominantly negative values. Conversely, the intercepts are confined to a narrower range, in particular between 0 and 2.

Fig.\ref{fig:RegressionAnalysis}a and Fig.\ref{fig:RegressionAnalysis}b illustrate five regression curves and their plots for a single output gene-perceptron, \textit{b1013}, at various timesteps. Fig.\ref{fig:RegressionAnalysis}a presents five  quadratic coefficients for the gene-perceptron \textit{b1013} under different weight configurations: 0 for $W_{10}$, -0.13 for $W_{40}$, 1.10 for $W_{70}$, 1.17 for $W_{100}$, and 1.17 for $W_{130}$. Notably, under the $W_{10}$ configuration, gene-perceptron \textit{b1013} exhibits a linear function, while the subsequent configurations result in regression lines with increasing curvatures. These results show how temporal plasticity can expand the solution space. Our previous study \cite{Somathilaka2023}, presented regressions for output gene-perceptrons based on a static weight configuration, allowing only a single regression function per output gene-perceptron. However, with the introduction of temporal plasticity weights, numerous regression curves can be derived for output gene-perceptrons.

\begin{figure}
    \centering
    \includegraphics[trim={10 20 0 10},clip,width=0.90\linewidth]{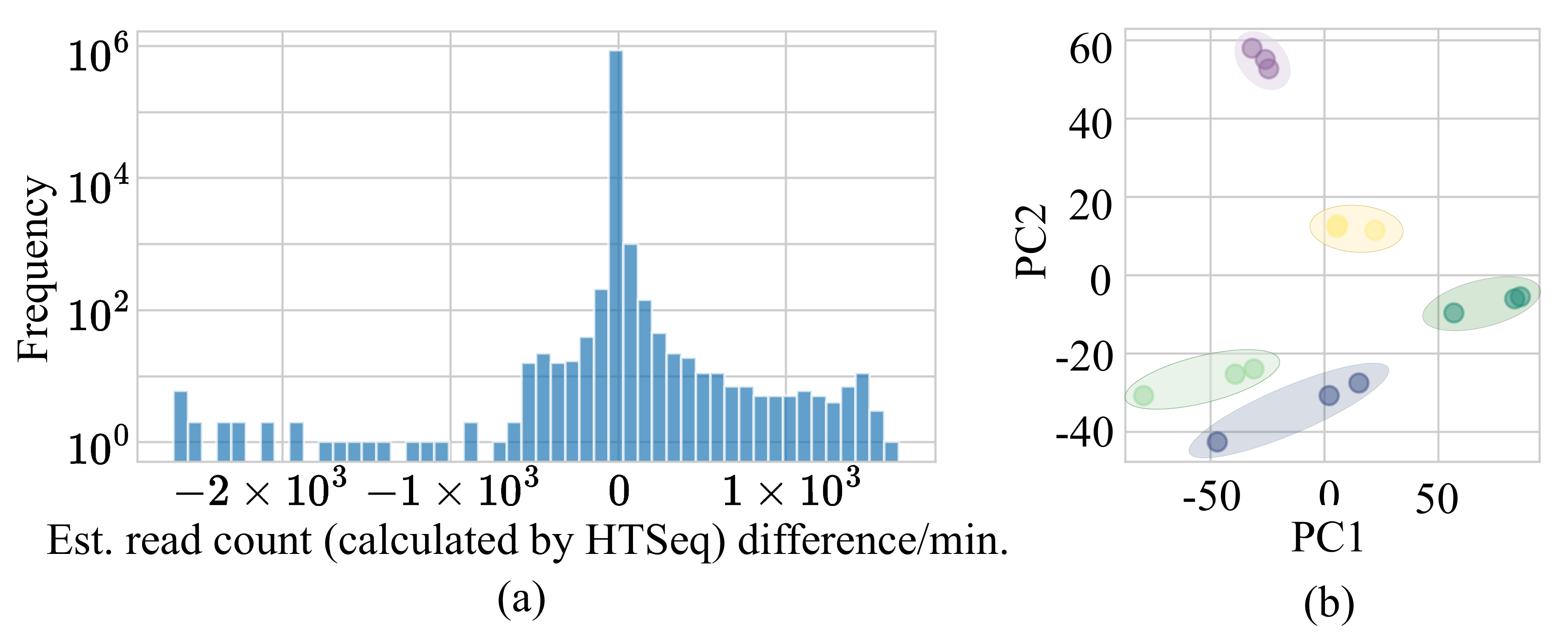}\vspace{-1.2em}
    \caption{Illustrations of a) estimated per-minute gene expression differences, and b) PCA clustering of experimental replicates, showing that similar environmental setups have close expression patterns, reinforcing reliable computing.}\vspace{-2.2em}
    \label{fig:Speed_and_Reliability}
\end{figure}

It is important to note that the evolution of the regression curve influenced by weight plasticity in GRNN computing is gradual. This means that systems requiring static weights can only operate in a certain period to guarantee stable computing results. Additionally, an analysis of read count change showed that cells can achieve a maximum value of -2,649.76 as shown in Fig. \ref{fig:Speed_and_Reliability}a, ensuring significantly fast computing for a biological entity.
We further conduct a principal component analysis of expression levels from temporal experiment replicates as shown in Fig. \ref{fig:Speed_and_Reliability}b, where each cluster of replicates, consistent across different conditions, confirms this reliability. Additionally, the concept of temporal plasticity offers a promising approach for addressing dynamic systems, a focus for our future research.\vspace{-1em}

\section{Conclusion}
\label{sec:conclusion}
Consistent with the vision of TinyML to establish miniature machine learning algorithms that can fit into low-powered devices, we extend our previously introduced GRNN concept towards a new paradigm with chemical-based ML algorithms found in biological cells. This new paradigm called Wet TinyML will transform a gene regulatory process into a GRNN that can compute similarly to a conventional ANN. Using the concept of bacterial cell plasticity, we show that the weights of the GRNN can be modified, opening new opportunities to map to diverse applications. Simultaneously, we estimate the energy consumption of GRNN subnetworks and find that they use less energy compared to traditional Von Neumann and Neuromorphic platforms. Our future research will evaluate the impact of noise in the GRN, cell reusability based on plasticity,  computing speed analyzed from wet-lab experiments. The wet lab experiments will explore input genes that can easily be stimulated and engineering reporter genes in the output layer to determine the correct GRNN computing in the subnetwork. The concept of Wet TinyML can expand the paradigm of miniature machine learning algorithms based on using its natural chemical reactions, which will take Biocomputing to a new level. This can result in new healthcare implantables that embed  engineered cells or Bio-hybrid computing systems, where computation is performed in  synergy between the biological cells and silicon technologies.\vspace{-1em}

\section{Acknowledgments}

\short{This work was supported by Science Foundation Ireland (SFI) and the Department of Agriculture, Food and Marine on behalf of the Government of Ireland (Grant No. 16/RC/3835), and the National Science Foundation (NSF) (Grant No. 2316960). Authors also like to thank Clare Lyle (Google DeepMind) for valuable input.}

\bibliographystyle{ACM-Reference-Format}
\bibliography{acmart}

\end{document}